\def\year{2022}\relax
\documentclass[letterpaper]{article} 
\usepackage{aaai22}  
\usepackage{times}  
\usepackage{helvet}  
\usepackage{courier}  
\usepackage[hyphens]{url}  
\usepackage{graphicx} 
\usepackage{natbib}  
\usepackage{caption} 
\DeclareCaptionStyle{ruled}{labelfont=normalfont,labelsep=colon,strut=off} 
\usepackage[linesnumbered,lined,ruled]{algorithm2e}
\usepackage{array}
\usepackage{amsmath}
\usepackage{amssymb}
\usepackage{comment}
\usepackage{subcaption}

\newcommand{\rif}{\stackrel{\,\,+}{\leftarrow}}

\newtheorem{example2}{\bf Example Domain}

\copyrighttext{Presented at the AI-HRI Symposium at AAAI Fall Symposium Series (FSS) 2022}

\title{Knowledge-based and Data-driven Reasoning and Learning for\\ Ad Hoc Teamwork}
\author{
    Hasra Dodampegama, Mohan Sridharan \\
}
\affiliations{

    School of Computer Science\\
    University of Birmingham, UK\\
    hhd968@student.bham.ac.uk, m.sridharan@bham.ac.uk
}

\begin{document}
\maketitle

\begin{abstract}
We present an architecture for \textit{ad hoc teamwork}, which refers to collaboration in a team of agents without prior coordination. State of the art methods for this problem often include a data-driven component that uses a long history of prior observations to model the behaviour of other agents (or agent types) and to determine the ad hoc agent's behavior. In many practical domains, it is challenging to find large training datasets, and necessary to understand and incrementally extend the existing models to account for changes in team composition or domain attributes. Our architecture combines the principles of knowledge-based and data-driven reasoning and learning. Specifically, we enable an ad hoc agent to perform non-monotonic logical reasoning with prior commonsense domain knowledge and incrementally-updated simple predictive models of other agents' behaviour. We use the benchmark simulated multiagent collaboration domain \emph{Fort Attack} to demonstrate that our architecture supports adaptation to unforeseen changes, incremental learning and revision of models of other agents' behaviour from limited samples, transparency in the ad hoc agent's decision making, and better performance than a data-driven baseline.
\end{abstract}

\section{Introduction}
\label{sec:introduction}
\begin{figure}[tb]
    \begin{center}
    \includegraphics[width=0.4\columnwidth]{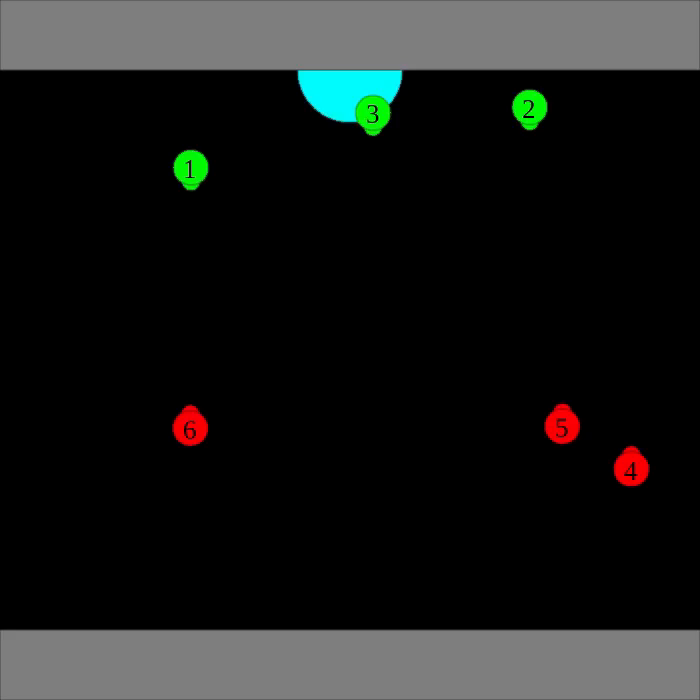}
    \includegraphics[width=0.4\columnwidth]{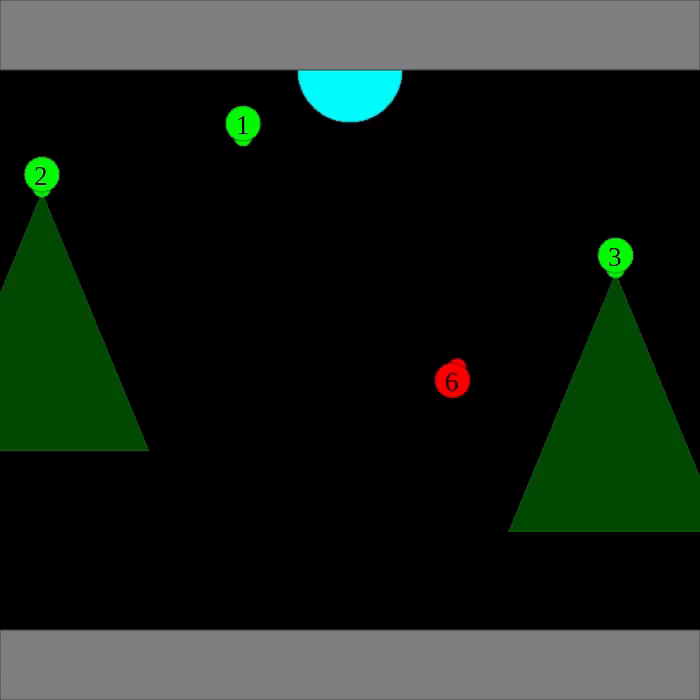}
    \vspace{-0.5em}
    \caption{Screenshots from \textit{fort attack} environment.}
    \label{fig:example}
    \end{center}
    \vspace{-1.2em}
\end{figure}

Ad hoc teamwork (AHT) refers to the problem of enabling an agent to cooperate with others on the fly~\cite{sk:10}. 
As a motivating example, consider \textit{Fort Attack}, a simulated benchmark for multiagent collaboration~\cite{ad:20}. Figure~\ref{fig:example} shows scenarios in which three guards (in green) are trying to protect a fort from three attackers (in red). One of the guards ("1") is the ad hoc agent, and an episode ends when all members of a team are terminated, an attacker reaches the fort, or guards manage to protect the fort for a sufficient time period. Each agent can move in a particular direction with a particular velocity, or shoot an opponent within a particular range. Although each agent is aware of the world state (e.g., location, status of each agent) at each step, individual agents have not worked with each other before and do not explicitly communicate with each other. This example poses challenging knowledge representation, reasoning, and learning problems. The ad hoc agent may have to reason with different (e.g., relational, probabilistic) descriptions of commonsense domain knowledge (e.g., default statements, domain attributes, and rules governing change) and uncertainty. The agent may also have to revise its knowledge and action choices in response to changes in domain (or agent) attributes and team composition. These problems are also representative of many other practical multiagent collaboration scenarios (e.g., disaster rescue, surveillance) that involve teams of robots and humans working to achieve a shared objective.

There has been considerable research in AHT under different names, as discussed in a recent survey~\cite{mr:22}. Initial methods encoded protocols for collaboration, with the ad hoc agent planning a sequence of options based on the state. State of the art methods include a key "data driven" component to learn probabilistic and/or deep network-based models or policies that estimate the behavior of other agents (or agent "types") or optimize the ad hoc agent's actions, based on a history of experiences. The focus on optimization makes it difficult for these methods to fully exploit the available domain knowledge. In a departure from existing work, our architecture draws on research in cognitive systems, moves away from data-driven optimization, and seeks to achieve reasoning with commonsense domain knowledge, incremental learning, and transparency in decision making. Specifically, the architecture:
\begin{itemize}
\item Performs non-monotonic logical reasoning with commonsense domain knowledge, enabling the ad hoc agent to plan and execute actions to achieve the desired goal, and to provide relational descriptions of its decisions;

\item Uses reasoning to guide rapid learning and revision of simple anticipatory models that satisfactorily mimic the behaviour of other agents from limited examples; and

\item Uses the principles of ecological rationality~\cite{gigerenzer:MMM16} to identify and use attributes and heuristics that support reliable and efficient reasoning and learning. 
\end{itemize}
We use Answer Set Prolog~\cite{gelfond:aibook14} for non-monotonic logical reasoning, and use the Fort Attack domain to demonstrate that our architecture supports reliable, efficient, and transparent reasoning, learning, and adaption.

\section{Related Work}
AHT has been defined as the challenge of creating an agent capable of collaborating with previously unknown teammates towards achieving a common goal~\cite{sk:10}. As described in a recent survey, research in AHT has existed under different names for at least 15 years~\cite{mr:22}. Early work encoded specific protocols (or plays) for different scenarios and enabled an agent to choose (or plan a sequence of) relevant protocol(s) based on the current state~\cite{Bowling:AAAI05}. Other work has used sampling-based methods such as Upper Confidence bounds for Trees (UCT) for determining the ad hoc agent's action selection policy~\cite{ba:13}. UCT has been combined with biased adaptive play to provide an online planning algorithm for ad hoc teamwork~\cite{Wu:AAAI11}, and researchers have explored using Value Iteration or UCT depending on the state observations~\cite{ba:11}. Much of the current (and recent) research has formulated an ad hoc agent's action choices using policy methods, assuming an underlying Markov decision process (MDP) or a partially observable MDP (POMDP)~\cite{ba:17,ch:20,Santos:21,ra:21}. This has included learning different policies for different known teammate \textit{types} using Fitted Q Iteration and selecting between these policies depending on the teammate types seen during execution~\cite{ba:17}. Later work has also used attention-based recurrent neural networks to avoid switching between policies for different teammate types~\cite{ch:20}.

A key component of state of the art approaches for AHT learns to predict the behaviours of other agents by using probabilistic or deep neural network methods and a long history of prior interactions with similar agents or agent types~\cite{ba:11,ba:17,ra:21}; predictions of other agents' actions are used to optimise the ad hoc agent's actions. Different ideas have been introduced to make such learning more tractable. For example, recent work has used sequential and hierarchical variational auto-encoders to model beliefs over other agents, and meta-learned approximate belief inference and Bayes-optimal behaviour for a given prior~\cite{zintgraf:21}. Learned policy methods have been combined with adversarial teammate prediction to account for behaviour changing agents~\cite{Santos:21}, and a Convolutional Neural Network-based change point detection method has been developed to detect and adapt to changing teammate types~\cite{mr:19}. Other work has used the observation of current teammates and learned teammate models to learn a new model~\cite{ba:17}. Despite these innovative ideas, learning of models requires considerable time, computation, and training examples. Moreover, the internal mechanisms governing decisions based on the learned knowledge are opaque.




In a departure from existing work, our AHT architecture combines knowledge-based and data-driven reasoning and learning to enable the ad hoc agent to adapt to different agent behaviours and team compositions, and to provide transparency in decision making, as described below.


\section{Architecture for Ad Hoc Teamwork}
\label{sec:arch}
\begin{figure}[tb]
    \begin{center}
    \includegraphics[width=0.9\columnwidth]{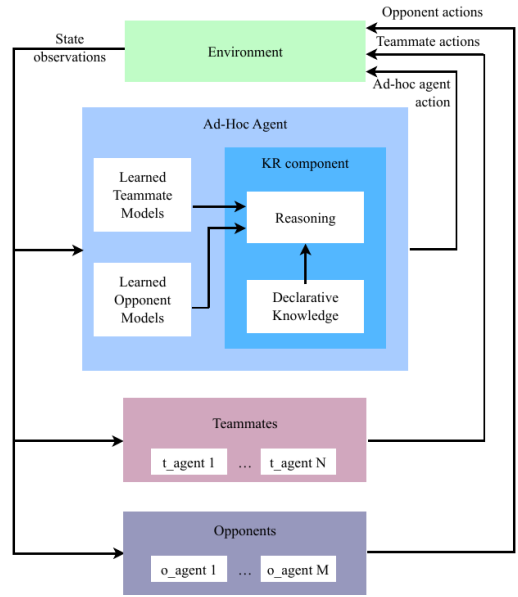}
    \vspace{-1em}
    \caption{Architecture combining knowledge-based and data-driven heuristic reasoning and learning.}
    \label{fig:arch}
    \end{center}
    \vspace{-1.1em}
\end{figure}

Figure~\ref{fig:arch} is an overview of the key components of our architecture. The ad hoc agent performs non-monotonic logical reasoning with prior commonsense domain knowledge and models of other agents' behaviours learned incrementally from limited examples, using heuristics to guide reasoning and learning. The individual components of our architecture are described using the following running example.

\begin{example2}\label{ex:illus-example}[Fort Attack (FA) Domain]\\
  {\rm Consider an instance of the fort attack domain with three guards protecting a fort from three attackers (Figure~\ref{fig:example}). One of the guards is our ad hoc agent that is able to adapt to changes in the team and domain. Prior commonsense domain knowledge includes relational descriptions of some domain attributes (e.g, location of agents), agent attributes (e.g., shooting capability and range), default statements (e.g., attackers typically spread and approach the fort), and some axioms governing change such as:
    \begin{itemize}
    \item An agent can only shoot others within its shooting range.
    \item An agent can only move to a location nearby.
    \end{itemize}
    This knowledge may need to be revised, e.g., some axioms and attribute values may change over time. }
\end{example2}

\subsection{Knowledge Representation and Reasoning}
\label{sec:arch-krr}
In our architecture, any domain's transition diagram is described in an extension of the action language $\mathcal{AL}_d$~\cite{gelfond:ANCL13} that supports non-Boolean fluents and non-deterministic causal laws~\cite{mohan:JAIR19}. Action languages are formal models of parts of natural language for describing transition diagrams of dynamic systems. $\mathcal{AL}_d$  has a sorted signature with \emph{actions}, \emph{statics}, i.e., domain attributes whose values cannot be changed by actions, and \emph{fluents}, i.e., attributes whose values can be changed by actions. \emph{Basic/inertial} fluents obey laws of inertia and can be changed  by actions, whereas \emph{defined} fluents do not obey laws of inertia and are not changed directly by actions. $\mathcal{AL}_d$ supports three types of statements: \textit{causal law}, \textit{state constraint} and \textit{executablility condition}; we will provide some examples of these statements later in this section.

\paragraph{Knowledge representation:} In our architecture, any domain's representation comprises system description $\mathcal{D}$, a collection of statements of $\mathcal{AL}_d$, and history $\mathcal{H}$. $\mathcal{D}$ has a sorted signature $\Sigma$ and axioms describing the transition diagram. $\Sigma$ defines the basic sorts, and describes domain attributes (statics, fluents) and actions in terms of the sorts of their arguments. For example, the basic sorts of FA domain include $ah\_agent$, $ext\_agent$, $agent$, $guard$, $attacker$, $dir$, $x\_val$, $y\_val$, and $step$ for temporal reasoning. These sorts are arranged hierarchically, e.g., the $ah\_agent$ and $ext\_agent$ are subsorts of $agent$. Statics of the FA domain include relations such as $next\_to(x\_val, y\_val, x\_val, y\_val)$ and $next\_dir(dir, dir)$, which encode the relative arrangement of places and directions. Fluents of the domain include:
\begin{align*}
   &in(ah\_agent, x\_val, y\_val), ~face(ah\_agent, dir),\\
   &shot(agent), ~agent\_in(ext\_agent, x\_val, y\_val),\\
   &agent\_face(ext\_agent, dir), ~agent\_shot(ext\_agent)
\end{align*}
which encode the location, orientation, and status of the ad hoc agent and other agents. Note that the last three relations are defined fluents. Domain actions include:
\begin{align*}
&move(ah\_agent, x\_val, y\_val), ~rotate(ah\_agent, dir),\\
&shoot(agent, agent), ~agent\_move(ext\_agent, x\_val, y\_val)\\
&agent\_rotate(ext\_agent, dir)
\end{align*}
which describe an agent's ability to move, rotate, and shoot, and to mentally simulate similar (exogenous) actions performed by other agents. $\Sigma$ also includes relations $holds(fluent, step)$ and $occurs(action, step)$ to imply that a particular fluent is true and a particular action occurs as part of a plan (respectively) at a particular step. Given this $\Sigma$, the axioms encoding the FA domain's dynamics include statements such as:
\begin{subequations}
\begin{align*}
    &move(R,X,Y) ~\mathbf{ causes }~ in(R,X,Y) \\
     &rotate(R,D) ~\mathbf{ causes }~ face(R,D) \\
    &\neg in(R,X1,Y1) ~\mathbf{ if }~ in(R,X2,Y2), ~X1 \neq X2, Y1 \neq Y2\\
    &\mathbf{impossible }~ shoot(R,A) ~\mathbf{ if }~ agent\_shot(A)
\end{align*}
\end{subequations}
which encode the expected outcome of an ad hoc agent's actions (i.e., move, rotate), specify that an agent cannot be in two locations at the same time, and prevent the consideration of an action whose outcome has been accomplished. The first two axioms are causal laws, followed by a state constraint and an executablility condition respectively.

The history $\mathcal{H}_c$ of a dynamic domain is usually a record of observations of basic fluents, i.e., $obs(fluent, boolean, step)$, and action executions, i.e., $hpd(action, step)$ at particular time steps. We also include default statements describing the values of fluents in the initial state in all but some exceptional situations, e.g., attackers usually spread and attack the fort, and they usually do not have the ability to shoot.
\begin{subequations}
\begin{align*}
    &{\bf initial\ default}\ spread\_attack(X)\ ~{\bf if}~~
        attacker(X)\\
    &{\bf initial\ default}\ \neg shoot(X, ah\_agent)\ ~{\bf if}~~
        attacker(X)    
\end{align*}
\end{subequations}
This representation does not assign numerical values to degrees of belief in these defaults, but supports elegant reasoning with the related beliefs and any specific exceptions.

\paragraph{Reasoning:} Key reasoning tasks of an agent include planning, diagnostics, and inference. For the ad hoc agent to reason with domain knowledge, the $\mathcal{AL}_d$ description is translated automatically to program $\Pi(\mathcal{D}_c, \mathcal{H}_c)$ in CR-Prolog, a variant of ASP that supports consistency restoring (CR) rules~\cite{balduccini:aaaisymp03}. ASP is based on stable model semantics and can represent recursive relations and constructs difficult to express in classical logic formalisms. It encodes \emph{default negation} and \emph{epistemic disjunction}, i.e., unlike ``$\lnot a$'' that states \emph{a is believed to be false}, ``$not~a$'' only implies \emph{a is not believed to be true}, and unlike ``$p~\lor\,\,\lnot p$'' in propositional logic, ``$p~or~\lnot p$'' is not tautologous. Each literal is thus true, false, or ``unknown'', and the agent only believes that which it is forced to believe. ASP also supports non-monotonic reasoning, i.e., revision of previously held conclusions, an essential ability in dynamic domains.


The program $\Pi(\mathcal{D}, \mathcal{H})$ for the FA domain has the signature and axioms of $\mathcal{D}$, inertia axioms, awareness axioms, reality checks, closed world assumptions for defined fluents and actions, observations, actions, and defaults from $\mathcal{H}$, and a CR rule for every default allowing the agent to assume  the default's conclusion is false to restore consistency under exceptional circumstances. For example, the CR rule:
\begin{align*}
  \lnot spread\_attack(X) \rif attacker(X)
\end{align*}
allows the ad hoc guard agent to consider the relatively rare situation of attackers mounting a frontal attack. Given a program $\Pi(\mathcal{D}, \mathcal{H})$, all reasoning tasks can be reduced to computing \emph{answer sets} of $\Pi$ after including suitable helper axioms. For example, helper axioms would be introduced to guide an agent to search for plans to achieve a goal. Our ad hoc agent can pursue different goals, e.g., shoot a nearby attacker or move to cover an unprotected region:
\begin{align*}
    &goal(I) \leftarrow holds(agent\_shot(attacker_2),I).\\
    &goal(I) \leftarrow holds(in(ag_{ah},X,Y),I), holds(spread\_guards,I). 
\end{align*}
The specific goal is set automatically by the ad hoc agent to maximise safety of the fort (see Algorithm~\ref{alg:adhoc-action}). Each answer set of $\Pi$, typically computed using a satisfiability solver, represents the ad hoc agent's beliefs about the world. It is also possible to encode heuristics that guide the computation of answer sets, e.g., to minimise the cost of the actions executed. We compute answer sets using the SPARC system~\cite{balai:lpnmr13}. Example SPARC programs for the FA domain are available in our repository~\cite{code-results}.


Knowledge-based reasoning paradigms are often criticized for requiring comprehensive prior knowledge in complex domains. However, ASP has been used by an international community to reason with incomplete knowledge, and modern ASP solvers are efficient for large knowledge bases~\cite{erdem:KI18}. The effort involved in encoding knowledge is typically much less than training purely data-driven systems. Also, as we have shown in other work in robotics~\cite{mota:SNCS21,mohan:JAIR19}, most of the knowledge about a domain only needs to be encoded once and can be revised over time.


\subsection{Agent Models and Heuristics}
\label{sec:arch-agent-models}
As stated earlier, a key component of an architecture for an ad hoc agent is the ability to learn models of other agents' behaviours. In the FA domain, the other agents include teammates (i.e., other guards) and opponents (i.e., the attackers). We seek to build simple predictive models that enable both rapid incremental updates and accurate predictions. To simulate different behaviours, we handcrafted four policies for different types of agents---two types each of guards and attackers---mimicking strategies used in the FA domain; the ad hoc agent is unaware of these types of agents. We used these agents in the FA domain and recorded details of states, observations, and action choices (e.g., 10K examples) during a few episodes executed in the domain. 

\paragraph{Ecological rationality and heuristics:} To build predictive models, we used the \textit{Ecological Rationality} (ER) approach, which builds on Herb Simon's original definition of \textit{Bounded Rationality}~\cite{gigerenzer:bookchap20} and the related rational theory of heuristics~\cite{gigerenzer:ARP11}. Unlike the focus on optimal search in many disciplines (e.g., finance, computing), ER focuses on the study of decision making under true uncertainty (i.e., in potentially open worlds), characterises behaviour (of agent or human) as a function of the internal (cognitive) processes and the environment, and focuses on \emph{satisficing}. Also, unlike the use of heuristics as a "hack", or to explain biases or irrational behaviour (e.g., of humans, in psychology), ER considers heuristics as a strategy to ignore part of the information in order to make decisions more quickly, frugally, and/or accurately than more complex methods~\cite{gigerenzer:ARP11}. It advocates an adaptive toolbox of classes of simple heuristics (e.g., one-reason, sequential search, lexicographic heuristics), and an algorithmic model involving the use of rigorous and comparative out-of-sample testing to identify heuristics that exploit the structure of the target domain. Such an approach has been shown to lead to very good performance in many practical applications~\cite{gigerenzer:MMM16} that are characterised by the environmental factors also observed in  AHT (discussed further below).

\newcolumntype{M}[1]{>{\centering\arraybackslash}m{#1}}
\newcolumntype{R}[1]{>{\raggedright\arraybackslash}m{#1}}
\begin{table}[tb]
\centering
\begin{tabular}{ | R{5.6cm} | M{1.2cm}| } 
  \hline
  \textbf{Description of attribute} & \textbf{Number}\\ 
  \hline
  x position of agent & 6 \\ 
  \hline
  y position of agent & 6 \\ 
  \hline
  distance from agent to center of field & 6 \\
  \hline
  agents' polar angle with center of field & 6 \\
  \hline
  orientation of the agent & 6 \\
  \hline
  distance from agent to fort & 6 \\
  \hline
  distance to nearest attacker from fort & 1 \\
  \hline
  number of attackers not alive & 1 \\
  \hline
  previous action of the agent & 1 \\
  \hline
\end{tabular}
\vspace{-0.5em}
\caption{Attributes considered by ad hoc agent to build simple predictive models of other agents' behaviour.}
\label{table:features}
\vspace{-1em}
\end{table}

\paragraph{Representational choices and models:} Drawing on the ER approach, we applied simple statistics to the small training set and identified the pose (i.e, position, orientation) of each agent and its recent action as the key factors defining behaviour, as listed in Table~\ref{table:features}. Since the values of these attributes range over a large span, we then built on the principles of abstraction and refinement in ER. In particular, we considered polar coordinates and relative distance of each agent from the fort. We also built on prior work by one of the authors on refinement in robotics~\cite{mohan:JAIR19} to describe and reason about positions at the level of coarser (abstract) regions and finer-granularity grid cell locations that are components of these regions. We formally coupled the two descriptions ($\mathcal{D}_C, \mathcal{D}_F$) through component relations and bridge axioms such as:
\begin{align*}
    &in^*(A, R) ~~\mathbf{if}~~ in(A, X, Y), ~component(X, Y, R)\\
    &next\_to^*(R_2, R_1) ~~\mathbf{if}~~ next\_to^*(R_1, R_2)
\end{align*}
where grid cell location $(X, Y)$ is in region $R$. 
This one-time formal encoding of the relationship between the two descriptions enabled the ad hoc agent to automatically choose the relevant part of the descriptions at run-time based on the target action, and to propagate information between the two granularities. For example, the location of an opponent farther away from the ad hoc agent can be in the form of regions (instead of grids) to simplify computation.

\begin{figure}[tb]
    \begin{center}
    \includegraphics[width=0.3\textwidth]{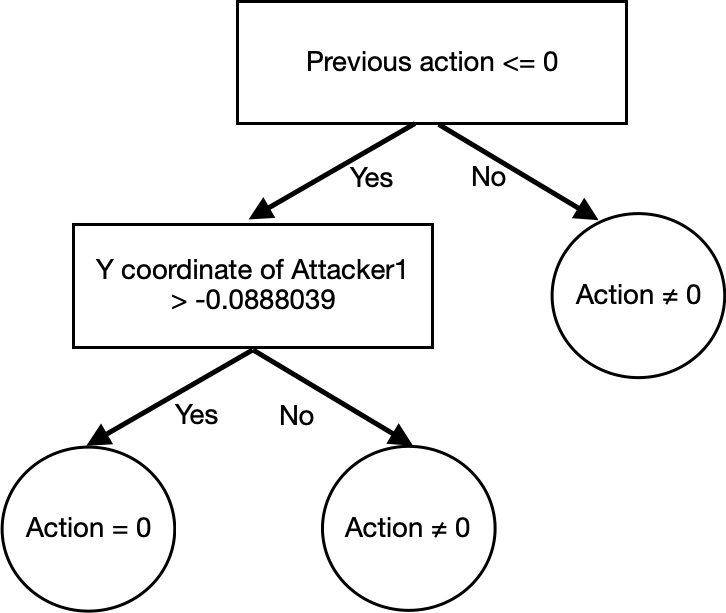}
    \vspace{-0.3em} 
    \caption{Fast and frugal tree for an attacker type agent.}
    \label{fig:tree}
    \end{center}
    \vspace{-1em}
\end{figure}

\begin{figure}[tb]
    \begin{center}
    \includegraphics[width=0.3\textwidth]{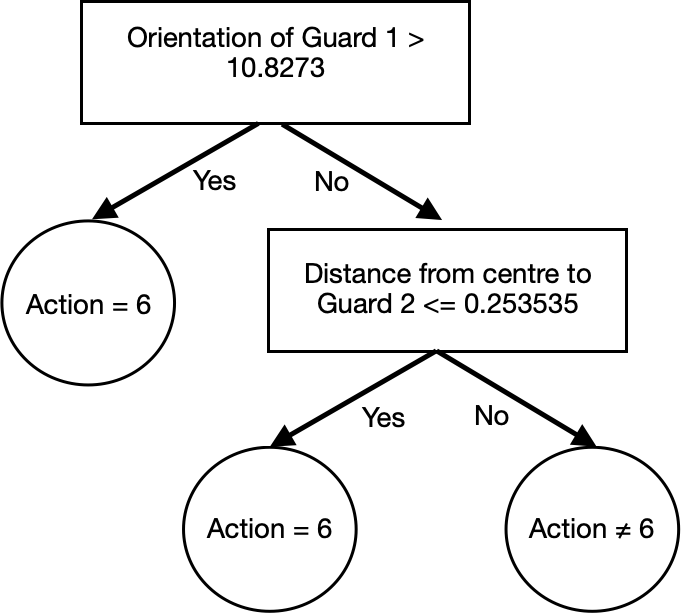}
    \vspace{-0.3em}
    \caption{Fast and frugal tree for a guard type agent.}
    \label{fig:tree_g}
    \end{center}
    \vspace{-2em}
\end{figure}

Once these representational choices are implemented, we matched the FA domain's environmental factors with the adaptive heuristics toolbox. In particular, we observed that the FA domain was characterised by dynamic changes, and the ad hoc agent had to revise its models (of other agents' behaviour) using limited samples and make decisions rapidly. We thus chose to explore: (i) an ensemble of "fast and frugal" (FF) decision trees in which each tree provides a binary class label and has its number of leaves limited by the number of attributes~\cite{gigerenzer:ARP11,katsikopoulos:book21}; and (ii) STEW, a regularised (linear) regression approach that exploits feature directions and is biased towards an equal weights solution~\cite{lichtenberg:icml19}.
We performed statistical testing on unseen examples, e.g., using ANOVA~\cite{Fisherbook:92}, to choose the FF trees-based models for behaviour prediction. 
More specifically, to build a model for a particular type of agent, we used a separate decision tree that mapped the state description to the binary choice of each action. We then trained another decision tree to map the output of these decision trees to the overall estimated action choice of the type of agent under consideration. Some individual FF trees learned for a particular attacker and guard are shown in Figures~\ref{fig:tree}-~\ref{fig:tree_g}.  

Unlike many existing methods, these predictive models can be learned and revised incrementally and rapidly. Also, consistent disagreement (agreement) with predictions of an existing model can trigger model revision (choice). For example, $\sum_{i=1}^N c_{i,j}/N$ can be computed for each of the $j$ agent types, where $c_{i, j} = 1$ when the predicted action of the $j^{th}$ behavioural model matches the actual action of the agent. If this fraction stays above a threshold for a particular model, that model continues to be used to predict the agent's next action; if it falls below a threshold for all known models, it may be time to acquire a new behavioural model.

\paragraph{Transparency in reasoning and learning:} Research shows that transparency, particularly the ability to answer causal, contrastive and counterfactual questions, plays a key role in human reasoning, and improves performance and acceptability of automated decision-making systems~\cite{anjomshoae:aamas19,fox:ijcaiwrkshp17,miller:AIJ19}. Knowledge-based reasoning and the use of predictive models makes it straightforward to achieve such transparency in our architecture. Any question about a particular action choice (or belief) can be answered by identifying each \textit{active} axiom that influences this action, i.e., has the corresponding action as the consequence and has all its antecedents satisfied in the corresponding answer set. This process can be repeated up to a relevant time step to identify a sequence of relevant literals and axioms that can then be used to construct relational descriptions that "explain" the desired decision (or belief). This component of our architecture uses existing natural language processing tools to parse questions, and uses a constrained vocabulary and templates to construct textual answers based on the identified literals and axioms. It also builds on prior work in our lab on generating such relational descriptions as explanations of an agent's decisions and beliefs~\cite{mota:SNCS21}. Since quantitative results may not fully illustrate the capabilities of our architecture, we provide some qualitative examples of this process in Section~\ref{sec:expres-trace}. 

\begin{algorithm}[tb]
\normalsize
\caption{\textbf{Control Loop of Architecture}}
\label{alg:control-loop}
\KwIn{N: number of games; $\Pi(\mathcal{D}, \mathcal{H})$: core ASP program, $\mathcal{M}$: behaviour models of other agents; $\mathcal{P}$: other agents' policies}
\KwOut{game\_stats: statistics of games}
Create environment, load $\mathcal{P}$, initialise environment\\
\For{$i = 0$ \KwTo $N-1$} {
    $\mathbf{s} \gets$ state of environment
    
    \While{$\lnot$ $game\_over(\mathbf{s})$}{
        $\mathbf{a}_o\gets$ other\_agents\_action($\mathbf{s}, \mathcal{P}$)
        
        $a_{ah} \gets$ adhoc\_agent\_action($\mathbf{s}, \Pi, \mathcal{M}$)  
        
        $\mathbf{a} = \mathbf{a}_o \cup a_{ah}$
        
        $\mathbf{s}'$ = execute($\mathbf{s}, \mathbf{a}$) 
        
        update\_models($\mathcal{M}$)
        
        \eIf{$game\_over(\mathbf{s'})$}{
            update(game\_stats)
            
            initialise environment        
        }{
            $\mathbf{s} = \mathbf{s'}$    
        }
    } 
} 
\textbf{return} game\_stats

\end{algorithm}

\begin{algorithm}[tb]
\caption{\textbf{adhoc\_agent\_action}}
\label{alg:adhoc-action}
\KwIn{$\mathbf{s}$, $\Pi(\mathcal{D}, \mathcal{H})$, $\mathcal{M}$} 
\KwOut{$a$}
\eIf{$alive(adhoc\_agent)$} {
    $\mathbf{a}_o \gets$  action\_predictions($\mathcal{M}$)\\
    
    $\mathbf{s}' \gets$ simulate\_effects($\mathbf{a}_o$) \\
    
    zones $\gets$ compute\_relevance($\mathbf{s}, \mathbf{a}_o, \mathbf{s}'$)\\

    ASP\_program $\gets$ construct\_program($\mathbf{s}, \Pi$, zones)\\

    answer\_set $\gets$ SPARC(ASP\_program)\\
    
    $a \gets$ next\_action(answer\_set)
} {
    $a$ $\gets$ do\_nothing
}

\textbf{return} $a$
\end{algorithm}

\subsection{Control Loop}
\label{sec:arch-control-loop}
The overall control loop of our architecture is described in Algorithm~\ref{alg:control-loop}. First, the FA game environment is set up, including the other agents' policies $\mathcal{P}$ (Line 1) that are unknown to the ad hoc agent, before each of the $N$ games (i.e., episodes) are played (Lines 2-17). In each game, the other agents' actions are identified based on current state and $\mathcal{P}$ (Line 5), and the ad hoc agent's action is computed by reasoning with domain knowledge and learned models ($\mathcal{M}$) of other agents' behaviours (Line 6; Algorithm~\ref{alg:adhoc-action}). The actions are executed in the simulated environment to receive the updated state (Line 8). The observed state can also be used to incrementally learn and update $\mathcal{M}$, e.g., when observations do not match predictions (Line 9). The updated state is used for the next step (Lines 13-15); this process continues until the game is over. The game's statistics are stored for analysis before moving to next game (Lines 10-12).

In Algorithm~\ref{alg:adhoc-action}, the selection of the ad hoc agent's action is only valid if this agent is alive (Lines 1-7). The ad hoc agent first uses the learned behaviour models ($\mathcal{M}$) of the other agents (guards, attackers) to predict their next action (Line 2). It simulates the effect of this action to obtain a likely next state (Line 3). It uses this information to compute the relevant regions (\emph{zones}) in the domain that need special attention (Line 4). This information, in turn, is used to automatically determine the regions and related axioms in $\mathcal{D}$ to be considered in the ASP program and the level of abstraction to be used (see Section~\ref{sec:arch-krr}, description of refinement in Section~\ref{sec:arch-agent-models}; Line 4). The relevant information is also used to automatically prioritise a goal (e.g., getting to a suitable region, shooting a specific attacker), and create and solve the ASP program (Line 5). The answer set obtained by solving this program provides the next action to be executed by the ad hoc agent (Line 7, returned to Algorithm~\ref{alg:control-loop}).


\section{Experimental setup and results}
\label{sec:expres}
We experimentally evaluated the following hypotheses:
\begin{itemize}
    \item \textbf{H1:} our architecture enables adaptation to different teammate and opponent types;
    \item \textbf{H2:} our architecture supports incremental learning of other agents' models that provide accurate predictions from limited examples; 
    \item \textbf{H3:} our architecture provides better performance than a state of the art data-driven system for AHT; and
    \item \textbf{H4:} our architecture supports the generation of relational descriptions of the ad hoc agent's decisions and belief.
\end{itemize}
We evaluated these hypotheses in \textit{fort attack}, a simulation-based benchmark domain for research in multiagent systems. Each episode (i.e., game) started with three guards and three attackers; our ad hoc agent was one of the guards. As stated earlier, the guards protected the fort while the attackers tried to reach the fort. An episode ended when all members of a team were terminated, an attacker reached the fort, or guards managed to protect the fort for a sufficient time period. Each agent can do nothing, move in one of the four cardinal directions, turn clockwise or counterclockwise, or shoot (eight actions). \textbf{Performance measures} included the number of steps in an episode, \% wins of a particular type of agent, and the accuracy of action choices.

\subsection{Experimental Setup}
\label{sec:expres-setup}
For initial training, we hand-crafted two sets of policies for the attackers and other guards: (\textbf{Policy1}) guards stay close to the fort and try to shoot attackers, while attackers spread and approach fort; (\textbf{Policy2}) both guards and attackers spread and shoot their opponents. We designed these policies to capture basic behavioural characteristics in the domain. To simulate training from limited examples, we used only $10000$ examples of state observations and agents' actions to train the ensemble of FF trees. We then tested the ad hoc agent in: (\textbf{Exp1}) when other agents followed the handcrafted policies; and (\textbf{Exp2}) when other agents were guided by the following built-in policies of the FA domain:
\begin{itemize}
    \item Policy 220: guards place themselves in front of the fort and shoot continuously; attackers try to approach the fort.
    \item Policy 650: guards stay near the fort and try to shoot nearby attackers; attackers try to sneak from all sides.
    \item Policy 1240: guards spread out, willing to move out a bit from the fort, and try to shoot when attackers are nearby; attackers try to sneak in from all sides.
    \item Policy 1600: guards spread, are willing to move further out from fort, and try to shoot nearby attackers; some attackers approach and shoot the guards, while others stay back and wait for a chance to reach the fort.
\end{itemize}
These policies represented a challenging choice since \textit{the ad hoc agent never observed others following these policies} although some of the basic skills appear in the handcrafted policies. We considered three baselines without our ad hoc agent: \textbf{Base1} in \textbf{Exp1} with agents following the hand-crafted policies; \textbf{Base2} in \textbf{Exp2} with agents following the built-in policies; and \textbf{GPL}, a state of the art AHT method based on graph (deep) neural networks~\cite{ra:21} in \textbf{Exp3} as an extension of \textbf{Exp2}. Each reported data point in the results below was the average of $100$ episodes. For GPL, we took the statistics from the supplementary material for the corresponding paper. We used these experiments to evaluate \textbf{H1-H3}; \textbf{H2} was also evaluated on a separate dataset. To better illustrate the ability to answer questions about decisions, we evaluated \textbf{H4} qualitatively instead of choosing from the different quantitative measures available in literature.




\subsection{Experimental results}
\label{sec:expres-results}
\begin{figure}
    \begin{subfigure}{0.49\columnwidth}
        \includegraphics[width=\linewidth]{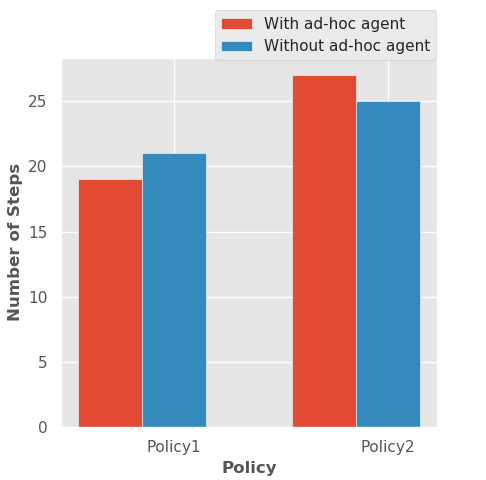}
        \caption{Average steps.}
        \label{fig:game_hc_a}
    \end{subfigure}
    \hspace*{\fill}
    \begin{subfigure}{0.49\columnwidth}
        \includegraphics[width=\linewidth]{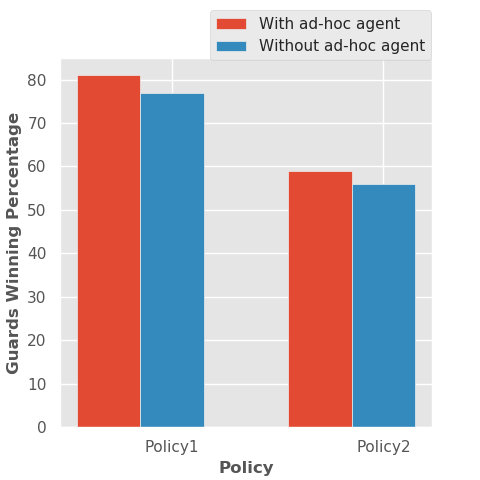}
        \caption{Average win \%.}
        \label{fig:game_hc_b}
    \end{subfigure}
    \caption{\textbf{(a)} Average number of steps in an episode with handcrafted policies and learned agent models. \textbf{(b)} Average \% of episodes in which guards win with handcrafted policies and learned agent models.}
    \label{fig:game_hc}
    \vspace{-1em}
\end{figure}

Figure~\ref{fig:game_hc} summarises the results of \textbf{Exp1}. When an ad hoc agent using the learned predictive models of other agents' behaviours was one of the guards, and other agents were guided by Policy1, the average number of steps in an episode was a little less than that with \textbf{Base1} that has all agents using the handcrafted Policy1---see Figure~\ref{fig:game_hc}(a). When the same experiment is run with handcrafted Policy2, the average number of steps in an episode is a little higher than that of the corresponding baseline. Figure~\ref{fig:game_hc}(b) shows that with Policy1 or Policy2, there was a (statistically) significant improvement in the average number of episodes in which the guards won, when our ad hoc agent was part of the team compared with the baselines. The difference in the results obtained with Policy1 and Policy2 in Figure~\ref{fig:game_hc}(a) can be attributed to the differences in the strategies followed by Policy1 and Policy2. Specifically, Policy2 is designed to favor the attackers, allowing them to launch a frontal assault on the guards. Hence, prolonging the duration of the game time while staying away from the attackers' shooting range is a favorable outcome for the guards with Policy2. Overall, the improvement in the \% wins observed in Figure~\ref{fig:game_hc}(b) is a good outcome of the use of our architecture.

\begin{figure}[t]
    \begin{center}
    \includegraphics[width=0.9\columnwidth]{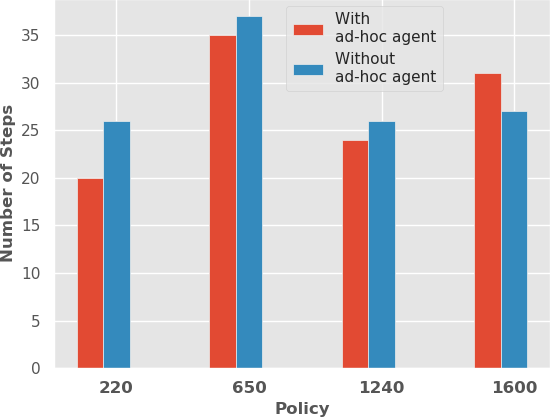}
    \caption{Average number of steps per episode with previously unseen built-in policies of FA domain; ad hoc guard agent reduces the number of steps (220, 650, 1240) or extends survival period (1600).}
    \label{fig:game_time_fa}
    \end{center}
    \vspace{-0.5em}
\end{figure}

\begin{figure}[t]
    \begin{center}
    \includegraphics[width=0.9\columnwidth]{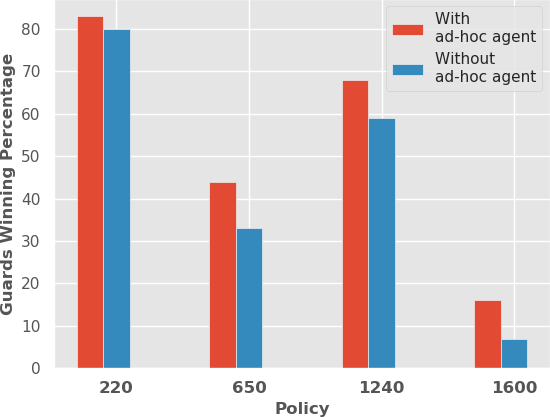}
    \caption{Average \% of episodes in which guards win when previously unseen built-in fort attack policies are used.}
    \label{fig:game_win_fa}
    \end{center}
    \vspace{-1.5em}
\end{figure}

Figures~\ref{fig:game_time_fa}-\ref{fig:game_win_fa} summarise the results of \textbf{Exp2}. We observed that the team of guards with an ad hoc agent using the behaviour models trained from the handcrafted policies was able to adapt to the previously unseen built-in policies of the FA domain. In particular, the average number of steps in an episode was lesser for policies 220, 650, and 1240, compared with \textbf{Base2} that has all agents using the built-in policies. At the same time, there was a statistically significant improvement in the \% wins for the team of guards with our architecture for these policies---see Figure~\ref{fig:game_win_fa}. The \% of episodes in which the team of guards won was rather small for policy 1600, although there is a significant improvement with our ad hoc agent ($16\%$ from $7\%$). Recall that this policy was particularly stacked against the guards, with some attackers trying to shoot the guards and draw them away from the fort while others stayed back and waited for an opportunity to sneak in. Notice, however, that the team with our ad hoc agent was able to prolong the game for much longer compared with \textbf{Base2}. These results support \textbf{H1} and provide partial support for \textbf{H2}.
They also show the benefits of having reasoning and learning inform and guide each other. \textit{Videos of some representative trials, including those corresponding to unexpected changes in team composition, are available in our open-source repository~\cite{code-results}}.

To further explore \textbf{H2}, we evaluated the behaviour models (trained as a stacked ensemble of FF trees from 10000 examples) on a previously unseen set of examples obtained using the handcrafted policies. The results summarised in Table~\ref{tab:accuracy} show that the prediction accuracy of these models is reasonably good in most cases, although there are still prediction errors. At the same time, the results discussed above indicated that this level of accuracy was sufficient to improve the performance of the team of guards. In other words, the behavioural models learned using the ER approach are able to strike a suitable balance between accuracy and simplicity, with the lack of over-fitting on the training set and the associated ability to revise models rapidly (potentially) being a key reason for performance improvement. These results provide further evidence in support of \textbf{H2}.


\begin{table}[tb]
\centering
\begin{tabular}{ | R{3cm} | M{1.5cm}| } 
  \hline
  \textbf{Agent Model} & \textbf{Accuracy} \\
  \hline
  \hline
  Guard type 1 & 85.47\%\\ 
  \hline
  Guard type 2 & 60.02\%\\
  \hline
  \hline
  Attacker type 1 & 86.89\%\\
  \hline
  Attacker type 2 & 85.22\%\\
  \hline
\end{tabular}
\vspace{-0.5em}
\caption{Prediction accuracy of learned behaviour models.}
\label{tab:accuracy}
\end{table}

\begin{figure}[t]
    \begin{center}
    \includegraphics[width=0.9\columnwidth]{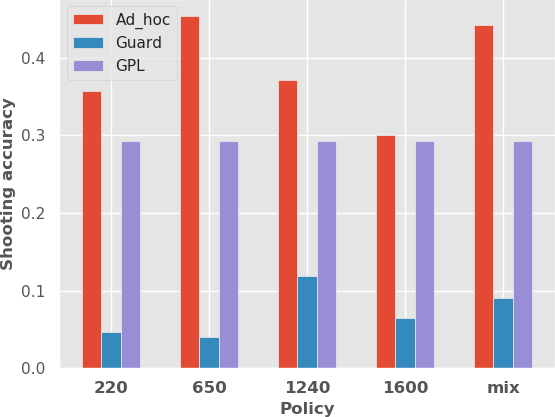}
    \vspace{-0.5em}
    \caption{Shooting accuracy for each built-in policy of FA domain. Our ad hoc guard agent has better accuracy compared with guard using baseline policies, including a state of the art data driven baseline.}
    \label{fig:shot}
    \end{center}
    \vspace{-1em}
\end{figure}

Figure~\ref{fig:shot} summarises the results of \textbf{Exp3} performed to evaluate \textbf{H3}, with the performance measure being \textit{shooting accuracy}, i.e., the fraction of times a guard agent's shooting eliminated an attacker. A high value of this measure would indicate a judicious use of the shooting capability and (more generally) the choice of actions that match the situation in the game. We compared our ad hoc guard agent with an ad hoc guard agent using GPL, and a guard using \textbf{Base2}, in the context of built-in domain policies. As stated earlier, the results for GPL were taken from their paper, which used shooting accuracy as a measure to show a marked improvement in performance compared with other data-driven (deep learning/RL) methods~\cite{ra:21}. 

We observed that our ad hoc guard agent's performance was much better than that of the ad hoc agent based on GPL in three policies (220, 650, 1240); results were comparable for the fourth (1600) policy. It is important to note that these policies represent challenging situations (650, 1240, 1600), e.g., guards are at high risk with policy 1600 because the attackers are trying to draw the guards out and shoot them. Also note that the GPL-based agent was trained for $60*160000$ steps to achieve the reported shooting accuracy, whereas our ad hoc agent reasoned with prior domain knowledge to avoid irrelevant actions and guide adaptation of the behaviour models for previously unseen policies. 

As an additional test, we considered agents (other than our ad hoc guard agent) using a randomly chosen combination of the built-in policies in each episode; the results are summarised by the final set of bars ("mix") in Figure~\ref{fig:shot}. Once again, we observed that our ad hoc agent had a much higher accuracy. These results demonstrated that the ad hoc agent shoots with higher accuracy compared with the other guards in the environment, regardless of the policy used by them. Note that reasoning allows the ad hoc agent to choose the most appropriate times to shoot, resulting in better action selection and high shooting accuracy than other agents. These results provide evidence in support of \textbf{H3}.

\begin{figure}[tb]
    \begin{center}
    \includegraphics[width=0.5\columnwidth]{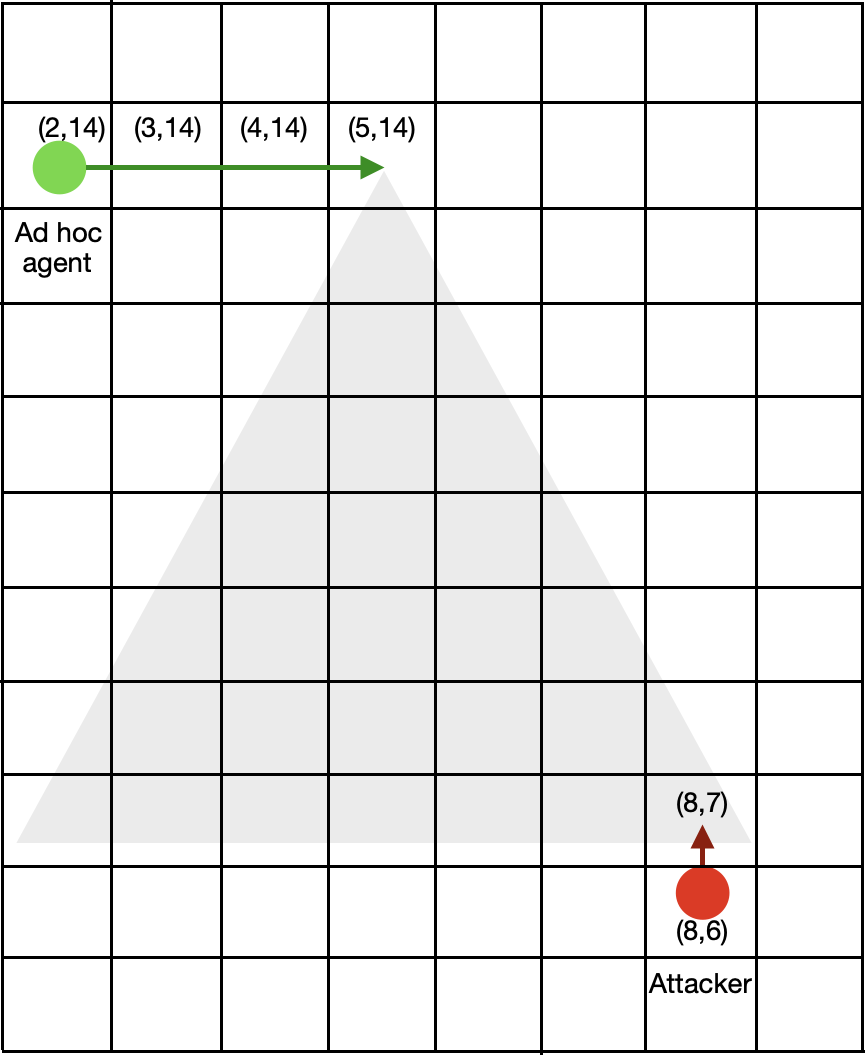}
    \caption{Part of the domain showing the ad hoc guard agent (green) moving to track and shoot an attacker (red).}
    \label{fig:explanation}
    \end{center}
    \vspace{-1.5em}
\end{figure}

\subsection{Execution trace}
\label{sec:expres-trace}
Consider an exchange with an ad hoc guard agent after an episode in which it tracked and shot an attacker; Figure~\ref{fig:explanation} shows a relevant subset of its actions and the domain.
\begin{itemize}
    \item \textbf{Human:} "Why did you move to (3,14) in step 1?"
    \item \textbf{Ad hoc Agent:} "Because attacker1 was not in range and I had to move to (4,14)". The answer presented the long(er)-term goal of getting attacker1 in range, and the associated short-term goal of getting to locations such as (4, 14) that will eventually enable it to shoot attacker1.
    \item \textbf{Human:} "Why did you not move to (5,13) in step 4?"
    \item \textbf{Ad hoc Agent:} "Because that would have put attacker1 out of range and I had to shoot attacker1"
    This answer demonstrates the ad hoc agent's ability to consider contrastive questions about actions it did not execute. This is an important ability for learning in humans.
\end{itemize}
Similar scenarios can be created for other types of questions, e.g., counterfactual ones. These results support \textbf{H4}.

\section{Conclusions and Future Work}
\label{sec:conclusions}
We described an architecture for ad hoc teamwork that combines knowledge-based and data-driven reasoning and learning. The architecture enables an ad hoc agent to perform non-monotonic logical reasoning with prior commonsense domain knowledge and fast and frugal decision tree models of the behaviour of other agents learned from limited examples. We demonstrated our architecture's capabilities in the benchmark fort attack domain, In particular, our ad hoc guard agent adapts to previously unseen teammates and opponents, rapidly revises the learned models, provides substantially better performance than baselines that include a state of the art data-driven method, and supports transparency in its decision making. 

The promising results provided by our architecture opens up multiple directions for further research that relax the limitations of our current work. The experiments reported in this paper focused on the FA domain with only one ad hoc (guard) agent in most of the experiments. Future work will explore more complex scenarios with multiple ad hoc agents in the FA domain and other benchmark multiagent collaboration domains. In addition, we will further explore how the interplay of reasoning and learning can be used to revise the behavioural models of other agents in response to dynamic changes. This interplay can also be used to consider when and how it would make sense for agents in a team to communicate relevant information with each other. Furthermore, we will explore the use of our architecture on physical robots in complex ad hoc teamwork settings.


\section*{Acknowledgments}
This work was supported in part by the US Office of Naval Research award N00014-20-1-2390. All conclusions are those of the authors alone. The authors thank Evgenii Balai for his help in making revisions to the SPARC implementation used in this paper.


\bibliography{aaai22}
\end{document}